\documentclass{beingbeyond}

\usepackage{colortbl}
\usepackage{tabularx}
\usepackage{array}
\usepackage{booktabs}
\usepackage{multirow}
\usepackage{algorithm}
\usepackage[noend]{algpseudocode}

\usepackage{amsmath,amssymb}
\usepackage{subcaption}
\usepackage[shortlabels,inline]{enumitem}
\usepackage{xspace}
\usepackage{adjustbox}
\usepackage[capitalize]{cleveref}

\DeclareRobustCommand{\onedot}{\futurelet\nexttoken\adddot}
\newcommand{\adddot}{\ifx\nexttoken.\else.\null\fi\xspace}
\newcommand{\eg}{\emph{e.g}\onedot}
\newcommand{\vs}{\emph{vs}\onedot}

\title{RA-RT: Robot Agents in Real Time via Learned Asynchronous Skill Scheduling}
\author{%
  \textbf{Chi Zhang}\textsuperscript{1}\thanks{Equal contribution.}\quad
  \textbf{Shumin Meng}\textsuperscript{2}\footnotemark[1]\quad
  \textbf{Yingge Xie}\textsuperscript{3}\footnotemark[1]\quad
  \textbf{Xiao Long}\textsuperscript{1}\\
  \textbf{Zhaoxi Chen}\textsuperscript{1}\quad
  \textbf{Wanpeng Zhang}\textsuperscript{2}\quad
  \textbf{Sipeng Zheng}\textsuperscript{2}\quad
  \textbf{Zongqing Lu}\textsuperscript{1,2}\thanks{Corresponding author.}\\[0.3em]
  \small\textsuperscript{1}Peking University\qquad
  \textsuperscript{2}BeingBeyond\qquad
  \textsuperscript{3}Zhejiang University\\[0.3em]
  \small\url{https://research.beingbeyond.com/ra-rt}%
}
\date{}

\begin{document}
\maketitle

\begin{abstract}
Large-model agents can tackle diverse manipulation tasks by composing reusable skills and adapting to execution feedback.
However, sequentially alternating planning and execution on robots typically wastes time and delays recovery from failures.
We propose \textbf{RA-RT}: Robot Agents in Real Time via Learned Asynchronous Skill Scheduling, a framework that coordinates perception, planning, motion, and recovery under a shared wall-clock budget.
During execution, the planner can commit follow-up skills, wait for fresh observations, or react to intermediate reports.
For real-time deployment, we train a compact planner through embodied fine-tuning and latency-aligned DAgger, pausing each student rollout while a slower teacher labels decisions from the student's observed states.
We further refine the planner on successful rollouts to encourage concise decisions and accelerate inference with output-aligned speculative decoding.
In simulation with the API-based teacher fixed, asynchronous scheduling improves completion over sequential execution across all three benchmarks within reasonable time budgets, and the full RA-RT system leads across most benchmarks and time budgets.
Our locally deployed 9B planner achieves higher completion rates in less time than prior baselines.
Without real-robot fine-tuning, it transfers to four single- and dual-arm real robot setups with low-latency execution.
\end{abstract}

\section{Introduction}
\begin{figure}[t]
    \centering
    \includegraphics[width=\textwidth]{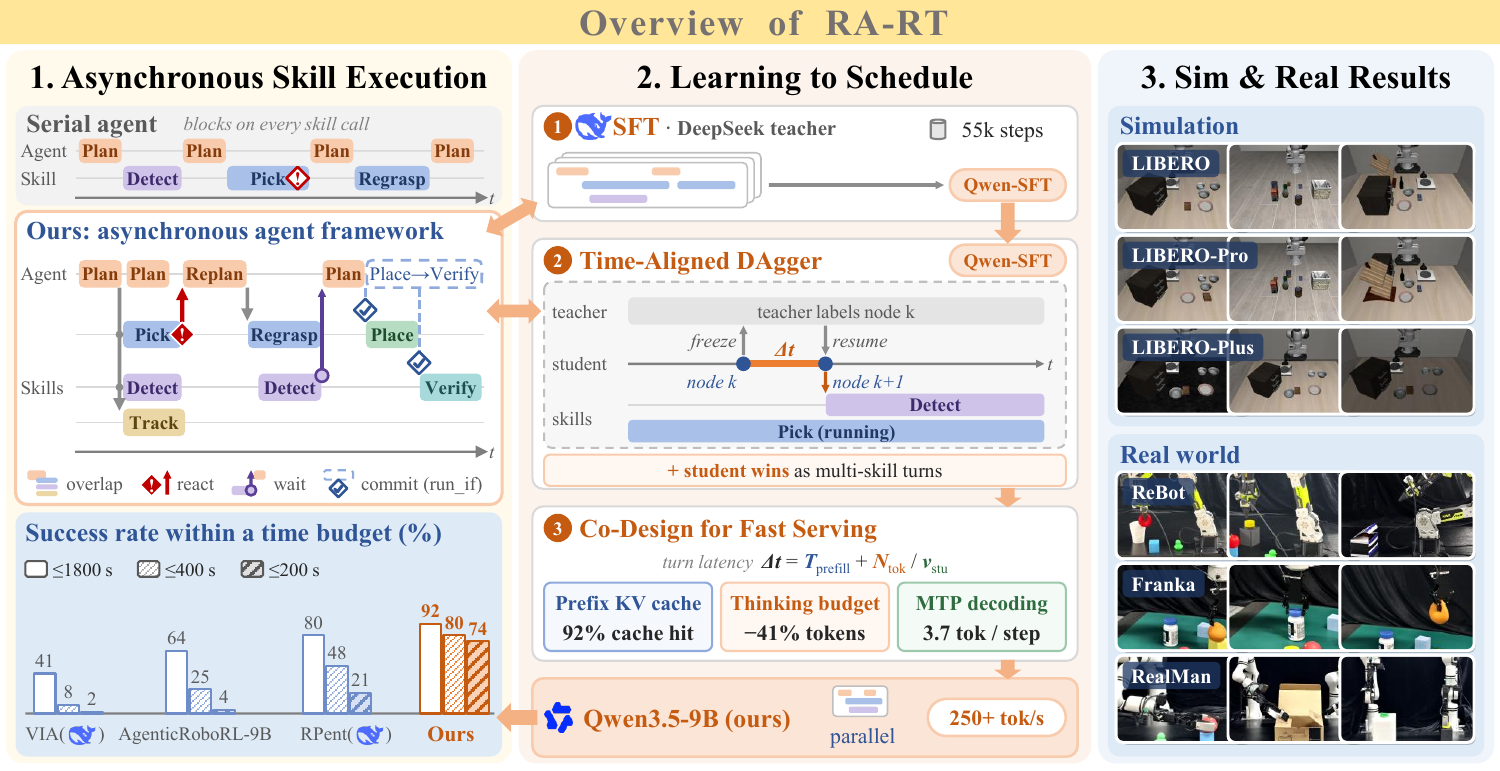}
    \caption{\textbf{Overview of asynchronous skill scheduling and compact planner training.}
    \textbf{Left:} Unlike blocking serial agents, our scheduler overlaps planning with execution, \emph{reacts} to mid-skill reports, \emph{waits} for post-motion observation when needed, and \emph{commits} conditional successor skills via \texttt{run\_if}, outperforming methods like VIA, AgenticRoboRL-9B, and RPent across time budgets.
    \textbf{Middle:} We fine-tune Qwen3.5-9B under DeepSeek-V4.1-Flash supervision, off-clock time-aligned DAgger, and multi-skill refinement, with low latency inference enabled by prefix KV caching, bounded reasoning tokens, and speculative MTP decoding.
    \textbf{Right:} Evaluation covers LIBERO, LIBERO-Pro, LIBERO-Plus, and physical ReBot, Franka, and RealMan robots.}
    \label{fig:teaser}
\end{figure}

Large models enable embodied agents to perform diverse robotic tasks by interpreting instructions, monitoring execution, and replanning after failures~\cite{saycan,inner,cap,harness,agp}.
These capabilities are particularly important for manipulation, where uncertainty and unexpected failures often demand recovery.
However, effective manipulation requires not only successful task completion but also timely execution.
Planning and retries introduce latency, while the physical world keeps changing.
We therefore study how embodied agents can schedule skills to efficiently coordinate perception, planning, and execution under a shared wall-clock budget.

Efficient scheduling requires deciding not only what to do, but also when to plan and act.
Consider a robot transporting a cup to a tray.
If the cup slips, the planner should react immediately rather than wait for the motion to finish.
If transport succeeds, the next camera observation can be scheduled in advance to avoid unnecessary planning delays.
Yet deciding where to place the cup requires fresh visual feedback and must wait until that observation becomes available.
An effective planner must therefore react early to failures, commit predictable actions ahead of time, and wait when new information is needed.

To enable these behaviors, we introduce an asynchronous skill-scheduling framework built on the RPent skill interface~\cite{harness} (\cref{fig:teaser}, left).
Rather than blocking until completion, each skill invocation immediately returns a \emph{pending receipt} indicating that execution is ongoing, allowing a single planner to reason concurrently with robot motion.
Intermediate execution reports provide timely feedback on progress and failures, enabling the planner to react before a skill completes.
Through \texttt{run\_if}, the planner can schedule follow-up skills that execute automatically once their predecessors satisfy specified conditions, while deferring decisions that require fresh observations.
Lightweight execution safeguards and read--write-separated dispatch ensure safe and consistent robot operation.

While the framework enables asynchronous scheduling, its effectiveness depends on the planner's decision latency, $\Delta t = T_{\mathrm{prefill}} + N_{\mathrm{tok}} / v_{\mathrm{stu}}$.
Unnecessary reasoning wastes time, while delayed decisions can miss opportunities for precommitment or recovery.
We therefore develop a three-stage learning and inference pipeline for a compact 9B planner (\cref{fig:teaser}, middle).
A key challenge is that the robot continues moving during inference, making the observed execution state dependent on planner latency.
For example, a slower teacher may observe a completed grasp that the fast student must still anticipate.
To bridge this gap, we first train the student through vision-frozen SFT on asynchronous trajectories, then query DeepSeek-V4.1-Flash~\cite{deepseekv41flash} \emph{off the robot clock} on saved student histories aligned to $\Delta t$~\cite{dagger}.
In summary, we make three contributions:

\begin{itemize}
\item \textbf{Asynchronous skill scheduling under a shared wall-clock budget.}
We formulate manipulation as a time-constrained planning problem and develop an asynchronous runtime featuring pending receipts, mid-skill interruption, outcome-gated \texttt{run\_if} precommitment, and physical transition guardrails.
This design enables planners to proactively commit predictable successor actions, wait for missing evidence, or react to failures during execution across different robots (\cref{sec:method,subsec:framework}).
\item \textbf{Low-latency planner training and inference.}
We train a compact 9B planner via supervised fine-tuning, time-aligned DAgger that labels paused student states without copying verbose teacher reasoning, and self-refinement on multi-skill rollouts.
At deployment, prefix KV caching, a reasoning token limit, and aligned speculative decoding keep per-turn latency low enough to act within motion windows (\cref{subsec:dataset,subsec:finetune,subsec:dagger,subsec:inference}).
\item \textbf{Real-time simulation and zero-shot real-robot validation.}
Across three real-time simulation benchmarks and four single- and dual-arm robot setups, asynchronous scheduling consistently improves completion across time budgets.
Without external APIs or real-robot fine-tuning, our local 9B planner outperforms blocking baselines and approaches the asynchronous teacher (\cref{sec:experiments}).
\end{itemize}

\section{Related Work}
\noindent\textbf{Vision-Language Models as Robot Planners.}
Vision-language and large language models increasingly serve as high-level robot planners, handling manipulation uncertainty by grounding instructions in feasible skills~\cite{saycan}, revising plans from execution feedback~\cite{inner}, or synthesizing control code and visual actions~\cite{capx,aspire,roborsi,via,inspectrobots}; in particular, RPent (Harness VLA)~\cite{harness} pairs a frozen VLA with perception and motion primitives, providing our simulation skill interface.
Most current planners, however, run in a strictly serial loop that stalls the robot after every skill---even when the next step needs no new observation---and delays recovery until motions finish.
On a continuously running physical clock, inference latency directly cuts into execution time~\cite{reflexfirst,gpt6astra}; even bounding single-call replanning cost via token budgets and latency targets, as in BRACE~\cite{brace}, cannot recover the wall-clock time lost while planning and motion wait on each other.

\noindent\textbf{Architectural Advances Beyond Sequential Execution.}
To break this serial bottleneck, recent studies have explored runtime architectures that cut planner round trips or overlap action selection with motion~\cite{concurrentcontrol}.
For predictable sequences, ChunkAct~\cite{chunkact} compiles task memory offline into fixed multi-primitive routines invoked in one call, trading step-by-step adaptability for fewer queries.
When downstream steps depend on runtime outcomes, non-blocking architectures intervene at different execution points: at skill boundaries, Retriever's manipulation agent adopts asynchronously revised plan suffixes~\cite{retriever} and REA routes fast policies through a boundary-triggered completion judge~\cite{rea}; during execution, DoReMi monitors skill constraints~\cite{doremi} and TypeGo interrupts active skills in a conditional stream~\cite{typego}.
These approaches differ in when a planner can revise an ongoing execution. We study when a successor can be committed using the information already available, and which conditions the executor must verify before dispatch. This distinction also matters during training because planner latency determines which execution reports are available at each decision.

\noindent\textbf{Coordinating Runtime Architecture and Planner Training.}
An asynchronous architecture pays off only when its planner knows which actions are safe to queue ahead and finishes reasoning within brief motion windows, requiring runtime design and model training to work together.
Existing planner post-training---whether distilling offline teacher trajectories~\cite{guava}, applying multi-turn RL to skill calls~\cite{agenticroborl}, or training simulation reasoning for zero-shot pose control as in PEAR~\cite{pear}---assumes a blocking loop where the world waits during inference.
Once skills keep running in an asynchronous runtime, the planner's latency determines which execution reports have arrived when it decides; even interactive imitation~\cite{dagger} and timing-aware expert takeover~\cite{timelydagger} leave a supervision mismatch if a slower remote teacher queried on a live clock observes later states than a fast local student.
Motivated by this coupling, \textbf{RA-RT} pairs an asynchronous scheduling architecture---decoupling argument-known \emph{commitment} by the planner from outcome-verified \emph{dispatch} by the executor under read--write consistency---with off-clock latency-aligned DAgger, concise multi-skill refinement, and aligned speculative local serving, improving wall-clock completion in real-time simulation and transferring zero-shot to single- and dual-arm robots via simulation training on the physical skill schema~\cite{agentflan,vctooler}.

\section{Method}
\label{sec:method}
\begin{figure}[t]
    \centering
    \includegraphics[width=\textwidth]{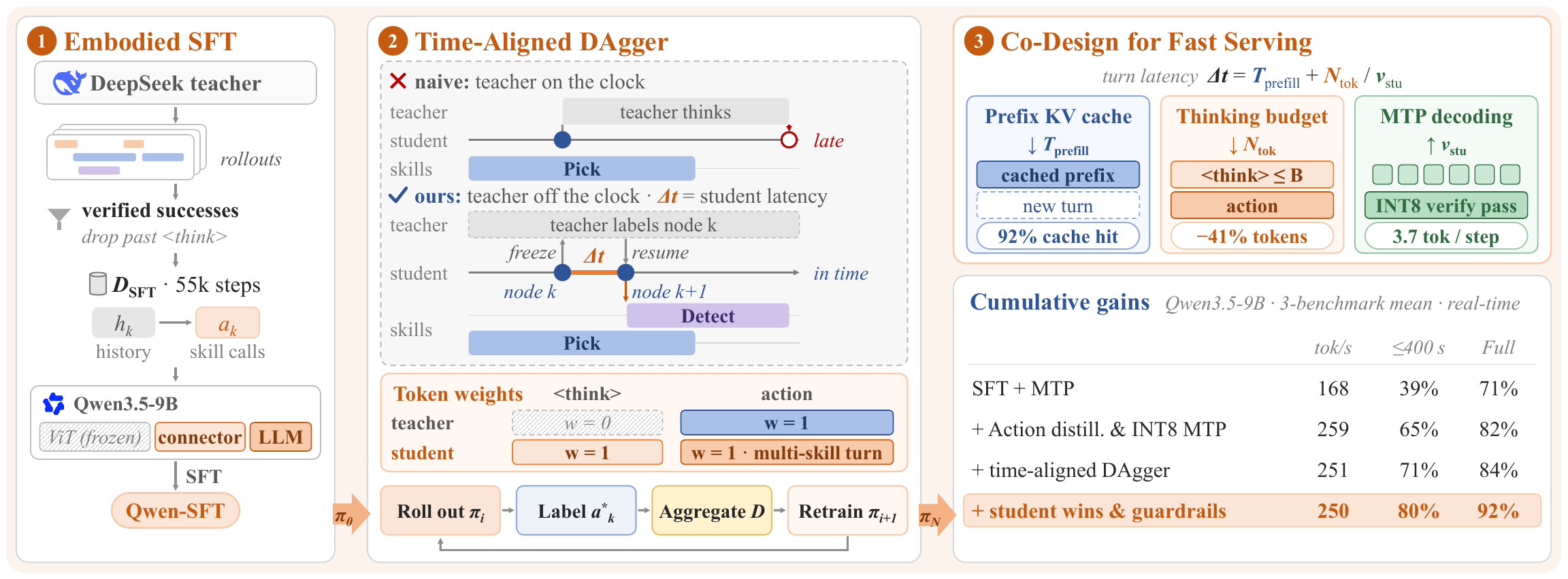}
    \caption{\textbf{Three-stage training and inference pipeline for low-latency skill scheduling.} \textbf{(1)} Vision-frozen SFT on verified asynchronous rollouts ($\mathcal{D}_{\mathrm{SFT}}$) from a DeepSeek-V4.1-Flash teacher trains the planner to map interaction history $h_k$ to skill calls $a_k$ while dropping past \texttt{<think>} blocks, producing $\pi_0$. \textbf{(2)} Off-clock time-aligned DAgger pauses student rollouts $\pi_i$ at step $k$, queries teacher action labels $a_k^*$ at student latency $\Delta t$, and retrains $\pi_{i+1}$ on $\mathcal{D}_{\mathrm{mix}}$ with teacher \texttt{<think>} masked ($w=0$) and multi-skill student rollouts retained ($w=1$). \textbf{(3)} Fast local inference reduces $\Delta t = T_{\mathrm{prefill}} + N_{\mathrm{tok}} / v_{\mathrm{stu}}$ through prefix KV caching, a reasoning token limit, and 8-bit speculative MTP decoding, substantially improving completion across tight and full budgets (\cref{tab:ablation_results}).}
    \label{fig:method_learning}
\end{figure}

We study robot manipulation on a continuous wall clock, where perception, model inference, motion, and recovery share a single time budget $B \le T_{\max}$ (with turn limit $N_{\max}$)~\cite{gpt6astra}. For a task instruction $I_{\mathrm{task}}$, let $T_{\pi}$ be the elapsed wall time until policy $\pi$ completes the task ($T_{\pi} = \infty$ on failure). We maximize the completion rate within budget $B$,
\begin{equation}
    F_{\pi}(B) = \Pr(T_{\pi} \le B),
    \label{eq:budget_objective}
\end{equation}
and minimize expected capped time $\mathbb{E}[\min(T_{\pi}, B)]$ at deployment budgets (\eg, $B \in \{200, 400, 800\}$\,s). At turn $k \in \{1, \dots, K\}$, assembled at wall time $t_k$, planner $\pi_{\theta}$ observes context
\begin{equation}
    x_k = \big(I_{\mathrm{task}},\, \mathcal{T},\, \mathcal{M},\, h_k,\, o_k\big),
    \label{eq:decision_context}
\end{equation}
consisting of the instruction $I_{\mathrm{task}}$, read-only and physical tool definitions $\mathcal{T} = \mathcal{T}_{\mathrm{ro}} \cup \mathcal{T}_{\mathrm{rw}}$, qualitative task notes $\mathcal{M}$, interaction history $h_k = \{(z_j, a_j, r_j)\}_{j=1}^{k-1}$ (past reasoning, tool calls, and delivered or pending receipts), and RGB-D plus proprioceptive observations $o_k$. Context compression $\tilde{x}_k = \mathcal{C}(x_k)$ removes past \texttt{<think>} blocks (\cref{subsec:dataset}) and condenses older frames and logs (\cref{subsec:inference}). Given $\tilde{x}_k$, the planner generates reasoning $z_k$ (\texttt{<think>}$\dots$\texttt{</think>}) and tool calls $a_k = (c_{k,1}, \dots, c_{k,m_k})$, denoted $y_k = (z_k, a_k)$. Each turn takes $\Delta t_k = T_{\mathrm{agent}}^{(k)} \approx T_{\mathrm{prefill}}^{(k)} + N_{\mathrm{tok}}^{(k)} / v_{\mathrm{stu}}$, where $T_{\mathrm{prefill}}^{(k)}$ is time-to-first-token (TTFT), $N_{\mathrm{tok}}^{(k)}$ is output token count, and $v_{\mathrm{stu}}$ is decoding speed. In a blocking loop, episode wall time sums
\begin{equation}
    T_{\mathrm{wall}} = T_{\mathrm{boot}} + T_{\mathrm{close}}
    + \sum_{k=1}^{K} \Big( T_{\mathrm{agent}}^{(k)} + T_{\mathrm{env}}^{(k)} + T_{\mathrm{glue}}^{(k)} \Big),
    \label{eq:wall_conservation}
\end{equation}
with setup/teardown $T_{\mathrm{boot}}, T_{\mathrm{close}}$, planner inference $T_{\mathrm{agent}}^{(k)}$ (context preparation, prefill, decoding), environment execution $T_{\mathrm{env}}^{(k)}$ (perception/VLA inference, real-time physics, capture), and runtime overhead $T_{\mathrm{glue}}^{(k)}$. Reducing $T_{\mathrm{wall}}$ combines (i) an \emph{asynchronous scheduling runtime} (\cref{subsec:framework}) overlapping $T_{\mathrm{agent}}^{(k)}$ with $T_{\mathrm{env}}^{(k)}$, and (ii) a \emph{training and inference pipeline} (\cref{fig:method_learning}; \cref{subsec:dataset,subsec:finetune,subsec:dagger,subsec:inference}) aligning supervision to $\Delta t_k$, shortening $N_{\mathrm{tok}}^{(k)}$, and raising $v_{\mathrm{stu}}$.

\subsection{Asynchronous Skill-Scheduling Framework}
\label{subsec:framework}

Built on the RPent interface~\cite{harness}, our runtime (\cref{fig:teaser}, left; \cref{alg:scheduler}) lets the planner reason during motion while keeping execution safe and state consistent.

\begin{algorithm}[!htb]
\caption{Asynchronous skill scheduling and guarded dispatch}
\label{alg:scheduler}
\footnotesize
\begin{algorithmic}[1]
\Require Instruction $I_{\mathrm{task}}$, budget $B$, write token $\nu \leftarrow 0$
\While{$t < B$ \textbf{and not} \texttt{terminated}}
    \State $(z_k, a_k) \leftarrow \pi_{\theta}(\mathcal{C}(x_k))$ \Comment{skills keep running}
    \For{each call $c = (\mathrm{op}, \alpha, p, \phi) \in a_k$} \Comment{omitted: \textbf{Wait}}
        \If{$c$ names a predecessor $p$} \Comment{\textbf{Commit}}
            \State Queue $c$ after $p$, gated by \texttt{run\_if} $\phi$
        \ElsIf{$\mathrm{op} \in \mathcal{T}_{\mathrm{ro}}$}
            \State Dispatch concurrently ($\le P$); stamp $(\nu, t_{\mathrm{obs}})$
        \Else
            \State Lock; $\nu \leftarrow \nu + 1$; launch $c$; return pending receipt
        \EndIf
    \EndFor
    \State Await a receipt or mid-skill report $e_t$; $k \leftarrow k + 1$ \Comment{\textbf{React}}
\EndWhile
\Statex \textbf{Executor}: on $p$ end, launch $c$ ($4$\,ms) if $\phi(r_p,q)\!=\!1$ and safe ($\nu{++}$ if $\mathcal{T}_{\mathrm{rw}}$), else skip; on fault $e_t$, abort, purge queue, $\nu{++}$, wake $\pi_{\theta}$.
\end{algorithmic}
\end{algorithm}

\noindent\textbf{Conditional Skill Commitment.}
Physical calls return pending receipts immediately, allowing $\pi_{\theta}$ to plan while the robot moves. Each call $c = (\mathrm{op}, \alpha, p, \phi)$ specifies arguments $\alpha$ and an optional predecessor $p$ with a \texttt{run\_if} condition $\phi$. Let $\sigma(x_k)$ be the information present in $x_k$, $\mathrm{dep}(\cdot)$ the inputs a quantity requires, and $(r_p, q)$ the predecessor's receipt and robot state when $p$ finishes. A follow-up call $c$ is \emph{safe to commit} at $t_k$ when its arguments are known and its condition depends only on executor-checkable outcomes:
\begin{equation}
    \mathrm{Safe}(c \mid x_k) \iff \mathrm{dep}(\alpha) \subseteq \sigma(x_k) \;\wedge\; \mathrm{dep}(\phi) \subseteq \{r_p,\, q\},
    \label{eq:admissible_commit}
\end{equation}
such as pre-grasp $\to$ grasp, transport $\to$ release $\to$ lift, or a post-arrival read queued after transport. Once $p$ ends, the executor checks $\phi$ locally and launches $c$ in $4$\,ms (versus $\sim\!1.4$\,s for a model call), or skips $c$ if $\phi$ fails. When $\alpha$ needs a future view (\eg, placement pose after arrival), $\pi_{\theta}$ \emph{waits} for $o_{k+1}$. On a mid-skill slip or stall ($e_t$), the executor aborts motion and purges queued \texttt{run\_if} calls before $\pi_{\theta}$ replans. Because the executor checks only $\phi$, $\pi_{\theta}$ decides when committing is valid.

\noindent\textbf{Skill Library and State-Consistent Execution.}
We retain each baseline's skill library and adopt RPent's~\cite{harness} for our simulation agent, including frozen LIBERO-tuned $\pi_{0.5}$~\cite{pi05,pi0} VLA skills, motion primitives, and SAM~3-based~\cite{sam3} 3D perception with depth. For real robots, we adapt the perception-and-motion skills to operate without simulator ground truth, support bimanual coordination absent from LIBERO, and use ReBot's wrist-only camera. We also add collision and accidental-contact safeguards absent from RPent's LIBERO skills. Full skill definitions and interface differences are provided in the supplement.
To keep state consistent, read-only perception ($\mathcal{T}_{\mathrm{ro}}$) runs concurrently without blocking (up to $P=8$; serialized on duplicate-object tasks), whereas each physical action ($\mathcal{T}_{\mathrm{rw}}$) acquires an exclusive lock and increments a write counter $\nu(t) \in \mathbb{N}$. Each read result $\rho$ records $\nu_{\rho}$ when requested; once a later action increments $\nu(t)$,
\begin{equation}
    \mathrm{Stale}(\rho, t) = \mathbb{I}\big[\nu_{\rho} < \nu(t)\big],
    \label{eq:stale_token}
\end{equation}
marks $\rho$ as outdated (\texttt{historical\_evidence}) so pre-motion object positions are not reused after motion, while in-flight reads record $t_{\mathrm{obs}}$ as transient until motion settles. On the bimanual setup, two planners share perception while each controls one arm without collision.

\noindent\textbf{Execution safeguards and episode restarts.}
Three simple checks prevent common failure loops: (i)~if the gripper closes on empty air ($\Delta q_{\mathrm{grip}} \le \delta_{\mathrm{air}}$), queued transport steps are canceled, and twisted wrists reset only beyond a safe tilt ($\theta_{\mathrm{tilt}} > \theta_{\max}$); (ii)~zoomed crop views are blocked from full-image depth projection; and (iii)~repeated static observations are capped, ending the episode as soon as the task succeeds. Each episode allows two or three clean-scene attempts: the first runs without notes, and later attempts reset the scene and attach task notes $\mathcal{M}$.

\subsection{Agentic Trajectory Dataset}
\label{subsec:dataset}

\begin{figure}[!htb]
    \centering
    \includegraphics[width=0.68\columnwidth]{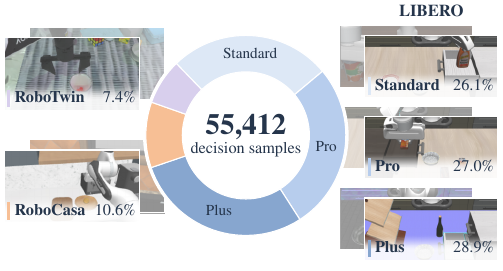}
    \caption{\textbf{Composition of the SFT dataset.} The dataset comprises 55,412 retained decision-level samples collected from LIBERO (Standard, Pro, and Plus), RoboCasa, and RoboTwin. Percentages show the share of samples per benchmark.}
    \label{fig:dataset_statistics}
\end{figure}

\noindent To run \cref{alg:scheduler} locally without cloud latency, we collect closed-loop teacher rollouts by running DeepSeek-V4.1-Flash~\cite{deepseekv41flash} inside our runtime across the LIBERO family~\cite{libero,liberopro,liberoplus}, RoboCasa~\cite{robocasa}, and RoboTwin~\cite{robotwin}.
Each trajectory is then unrolled into step-level supervision pairs $(\tilde{x}_k, y_k)$, where $y_k = (z_k, a_k)$ retains the current reasoning trace and skill calls, while $\tilde{x}_k = \mathcal{C}(x_k)$ strips \texttt{<think>} blocks from earlier turns ($j < k$).
Next, we measure token lengths over the formatted steps and drop any trajectory exceeding $64$K tokens in full rather than truncating it, avoiding partial episodes that cut off late-stage recovery and completion.
The surviving $2{,}593$ trajectories yield $\mathcal{D}_{\mathrm{SFT}}$ with $55{,}412$ decision samples (\cref{fig:dataset_statistics}), partitioned by $(\text{environment}, \text{task}, \text{seed})$ so validation and held-out evaluation initial states never appear in training.

Subsequent DAgger and refinement rollouts (\cref{subsec:dagger}) follow the same held-out test exclusions.
Finally, because post-trained VLM agents are sensitive to skill-schema changes~\cite{agentflan,vctooler,guava}, we construct a parallel dataset for our real-robot 9B planner by re-collecting teacher supervision with the physical-robot skill schema at the same simulation decision states across all five suites (supplement), matching the exact sample count and training recipe with zero real-robot data.

\subsection{Model Training}
\label{subsec:finetune}

\noindent\textbf{Supervised Fine-Tuning.}
Stage~1 (\cref{fig:method_learning}, left) fine-tunes Qwen3.5-9B~\cite{qwen35} on $\mathcal{D}_{\mathrm{SFT}}$ as our compact local planner; its hybrid Gated DeltaNet (GDN)~\cite{gateddeltanet} and full-attention backbone processes long $64$K-token interaction histories efficiently.
Keeping the vision encoder frozen to preserve visual representations and reduce memory overhead, we update the multimodal connector and LLM via next-token negative log-likelihood over both the current reasoning trace and skill calls $y = (z, a)$:
\begin{equation}
    \mathcal{L}_{\mathrm{SFT}}(\theta)
    = -\mathbb{E}_{(\tilde{x},y)\sim\mathcal{D}_{\mathrm{SFT}}}
    \left[ \sum_{t=1}^{|y|} \log p_{\theta}(y_t \mid \tilde{x}, y_{<t}) \right].
    \label{eq:sft_loss}
\end{equation}
The resulting policy $\pi_0$ provides a strong asynchronous baseline and initializes Stage~2.

\noindent\textbf{Time-Aligned DAgger and Self-Refinement.}
\phantomsection\label{subsec:dagger}
Offline SFT struggles when object layouts change and often imitates slow, single-step calls.
Stage~2 (\cref{fig:method_learning}, center) addresses both by rolling out the current student $\pi_i$ on disjoint training seeds $\mathcal{S}_{\mathrm{train}}$.
Let $\mathcal{E}(\xi, \Delta t)$ denote how the world state $\xi$ and running skills evolve over $\Delta t$. Querying the remote teacher ($\Delta t_{\mathrm{tch}} \gg \Delta t_k$) on a live $20$\,Hz clock lets the world drift for $\Delta t_{\mathrm{tch}}$ before its label takes effect. We instead pause simulation at $t_k$, query $a_k^* = \pi_{\mathrm{tch}}(\tilde{x}_k)$ \emph{off-clock}, and resume after the student's own latency $\Delta t_k$, aligning supervision to the deployment state:
\begin{equation}
    \xi_{k+1}^{\mathrm{on\text{-}clock}} = \mathcal{E}\!\big(\xi(t_k),\, \Delta t_{\mathrm{tch}}\big)
    \;\neq\;
    \mathcal{E}\!\big(\xi(t_k),\, \Delta t_k\big) = \xi_{k+1}^{\mathrm{aligned}}.
    \label{eq:time_aligned_state}
\end{equation}
Unlike post-hoc relabeling of unpaused student rollouts---where an invalid commit or missed $e_t$ enters unrecoverable states---executing $a_k^*$ after $\Delta t_k$ keeps rollouts recoverable while matching the student's in-flight view. These labels and teacher recovery rollouts form $\mathcal{D}_{\mathrm{DAg}}$ ($s=\mathrm{tch}$). We also collect successful real-time student rollouts $\mathcal{D}_{\mathrm{on}}$ ($s=\mathrm{stu}$), combining sequential motions into single-turn \texttt{run\_if} chains with shorter \texttt{<think>} blocks. Over $\mathcal{D}_{\mathrm{mix}} = \mathcal{D}_{\mathrm{SFT}} \cup \mathcal{D}_{\mathrm{DAg}} \cup \mathcal{D}_{\mathrm{on}}$ ($\mathcal{D}_{\mathrm{SFT}}$ tagged $s=\mathrm{tch}$), with each output split at $y_{\tau} = \texttt{</think>}$, we retrain $\pi_{i+1}$ via
\begin{equation}
\mathcal{L}_{\mathrm{Post}}(\theta)
= -\mathbb{E}_{(\tilde{x},y,s)\sim\mathcal{D}_{\mathrm{mix}}} \sum_{t=1}^{|y|} \bar{w}_t(s)
\log p_\theta(y_t \mid \tilde{x}, y_{<t}),
\label{eq:post_loss}
\end{equation}
where $\bar{w}_t(s) = w_t(s) / \sum_{j=1}^{|y|} w_j(s)$ uses weights
\begin{equation}
    w_t(s) =
    \begin{cases}
        1, & \text{if } s = \mathrm{stu}, \\
        0, & \text{if } s = \mathrm{tch} \text{ and } t \le \tau, \\
        1, & \text{if } s = \mathrm{tch} \text{ and } t > \tau.
    \end{cases}
    \label{eq:token_weights}
\end{equation}
Setting $w_t(\mathrm{tch})=0$ on $y_{1:\tau}$ lets the student learn teacher recovery actions $a_k^*$ without copying long teacher reasoning, while student rollouts keep reasoning concise.

\subsection{Model Merging and Fast Inference}
\label{subsec:inference}

Stage~2 checkpoints trade speed against recovery: multi-skill models excel at tight budgets, whereas DAgger models recover better on hard tasks.
Stage~3 (\cref{fig:method_learning}, right) averages their LoRA~\cite{lora} weights~\cite{modelsoups} ($\theta = \sum_{j} \beta_j \theta_j$, $\sum_j \beta_j = 1$) to combine both without extra runtime cost, then deploys $\theta$ with three optimizations that shrink $\Delta t_k = T_{\mathrm{prefill}}^{(k)} + N_{\mathrm{tok}}^{(k)} / v_{\mathrm{stu}}$.

\emph{Prefix-cached context} ($\downarrow T_{\mathrm{prefill}}^{(k)}$). To reuse the KV cache across long episodes, we keep only the latest $W_{\mathrm{img}}=8$ images (replacing older ones with placeholders) and, given $L$ messages, compress old tool logs in fixed blocks of $B_{\mathrm{msg}}=50$ while leaving the latest $L_{\mathrm{tail}}=25$ messages untouched:
\begin{equation}
    c(L) = \left\lfloor \frac{L - L_{\mathrm{tail}}}{B_{\mathrm{msg}}} \right\rfloor \cdot B_{\mathrm{msg}}
    \quad (B_{\mathrm{msg}}=50,\, L_{\mathrm{tail}}=25),
    \label{eq:block_compaction}
\end{equation}
so all tokens before $c(L)$ remain identical from turn to turn and hit the prefix KV cache.

\emph{Reasoning token limit} ($\downarrow N_{\mathrm{tok}}^{(k)}$). We cap each \texttt{<think>} block at $B_{\mathrm{think}}$ tokens; if reached, we inject \texttt{</think>} into the active sequence so the model outputs tool calls directly.

\emph{Aligned speculative decoding} ($\uparrow v_{\mathrm{stu}}$). We use Qwen3.5-9B's 6-step MTP~\cite{mtp} head $\psi$ for speculative decoding~\cite{specdecoding} and fine-tune it~\cite{fastmtp} on merged-student rollouts $\mathcal{D}_{\mathrm{stu}}$ ($\theta$ frozen):
\begin{equation}
\begin{aligned}
    \mathcal{L}_{\mathrm{MTP}}(\psi)
    &= \mathbb{E}_{(\tilde{x},y)\sim\mathcal{D}_{\mathrm{stu}}} \sum_{t} \sum_{m=1}^{M} \lambda_m \\[-2pt]
    &\quad D_{\mathrm{KL}}\!\Big( p_{\theta}(\cdot \mid \tilde{x}, y_{<t+m}) \,\Big\|\, q_{\psi}^{(m)}(\cdot \mid u_t^{\theta}, y_{t:t+m-1}) \Big),
\end{aligned}
    \label{eq:mtp_kl}
\end{equation}
using full-vocabulary forward KL via the shared frozen LM head ($u_t^{\theta}$, rollout tokens $y_{t:t+m-1}$, $\lambda_m\!=\!0.7^{m-1}$, $M\!=\!4$), with 6-step $(T, \text{top-}k, \text{top-}p)$ rejection sampling~\cite{specdecoding}.

\section{Experiments and Results}
\label{sec:experiments}

We  examine three questions: (1)~how much asynchronous scheduling improves task completion with a fixed DeepSeek-V4.1-Flash planner, (2)~how our compact Qwen3.5-9B planner compares with API-based and local baselines across time budgets, and (3)~how both planners transfer zero-shot to real robots.

\subsection{Experimental Setup}
\label{subsec:setup}

We evaluate on 40 LIBERO~\cite{libero}, 80 LIBERO-Pro~\cite{liberopro}, and 28 LIBERO-Plus~\cite{liberoplus} tasks in 20\,Hz real-time simulation (1800\,s wall-clock budget per task), reporting the unweighted mean success rate across benchmarks. Baselines include skill-based (RPent~\cite{harness}, AgenticRoboRL SFT/RL~\cite{agenticroborl}), code-generation and self-improving (CaP-Agent0~\cite{capx}, ASPIRE~\cite{aspire}, RoboRSI~\cite{roborsi}), and 3D visual agents (VIA~\cite{via}). Our agent shares RPent's skill-library structure. All methods use DeepSeek-V4.1-Flash~\cite{deepseekv41flash}, except AgenticRoboRL and our Qwen3.5-9B student, which run locally.
Each baseline starts from its native skill library and retains its planning-recovery loop without cross-task carryover, evaluated on one shared held-out initial state per task (\texttt{seed=1}) under uniform criteria (\cref{fig:tradeoff}: solved-task median; \cref{tab:ablation_results}: all-task \text{Med.}\ and $\mathrm{Cap.}=\mathbb{E}[\min(T_{\pi}, 1{,}800\,\text{s})]$). For CaP-Agent0 and ASPIRE, the budget includes exploration time; RoboRSI uses a frozen library without test-time updates. Ablations cap turns at 100 per task (150 across restarts for our final student; 360 for $1{,}800$\,s serial DeepSeek). RoboCasa~\cite{robocasa} and RoboTwin~\cite{robotwin} results are in the supplement due to limited open-source baseline support and follow the same trends.

\noindent\textbf{Real-world Experiments.}
We test zero-shot transfer across four setups (10 trials/pick and place, 5/knob or box, 60 total; \cref{fig:teaser}, right; \cref{fig:real_world}): single-arm Franka and RealMan (pick-and-place, knob turning), single-arm ReBot (adding box folding), and dual-arm RealMan (dual-knob turning, box folding). Franka-only RPent and Inspect Robots~\cite{inspectrobots} retain their native interfaces on DeepSeek-V4.1-Flash, while AgenticRoboRL (9B RL) and our serial and asynchronous variants use our hardware-safe skill set (\cref{subsec:framework,subsec:dataset}). Despite interface differences required for safety and sensing, all methods run strictly zero-shot without real-robot fine-tuning (local 9B models on a single RTX PRO 5000). Training and serving details are in the supplement.

\subsection{Main Results}
\label{subsec:main_results}

\noindent\textbf{Comparison with SoTA Baselines.}
\Cref{tab:benchmark_results} and \cref{fig:efficiency_analysis} compare full systems across wall-clock budgets. CaP-Agent0, ASPIRE, and RoboRSI achieve low completion under tight budgets, confirming that long-budget success does not translate into fast execution. While RPent leads among external baselines, RA-RT (DeepSeek) completes most tasks even under tight budgets and maintains its lead as the budget expands. Compared with RPent, which shares the same planner model and skill-library structure, RA-RT achieves higher completion rates with a shorter median completion time on successful episodes (\cref{fig:tradeoff}).

\begin{table}[!htbp]
\centering
\footnotesize
\caption{\textbf{Success rates (\%) under execution time budgets of 200, 400, and 800 seconds in real-time simulation ($1\times$ real-time clock).} Average is the unweighted mean across the three benchmarks. Model configurations and evaluation protocols are detailed in Sec.~4.1.}
\label{tab:benchmark_results}
\setlength{\tabcolsep}{4.5pt}
\renewcommand{\arraystretch}{1.1}
\begin{adjustbox}{max width=\linewidth}
\begin{tabular}{l*{4}{ccc}}
\toprule
\multirow{2}{*}{\textbf{Method}}
& \multicolumn{3}{c}{\textbf{LIBERO (40)}}
& \multicolumn{3}{c}{\textbf{LIBERO-Pro (80)}}
& \multicolumn{3}{c}{\textbf{LIBERO-Plus (28)}}
& \multicolumn{3}{c}{\textbf{Average}} \\
\cmidrule(lr){2-4}\cmidrule(lr){5-7}\cmidrule(lr){8-10}\cmidrule(lr){11-13}
 & 200\,s & 400\,s & 800\,s & 200\,s & 400\,s & 800\,s & 200\,s & 400\,s & 800\,s & 200\,s & 400\,s & 800\,s \\
\midrule
RPent~\cite{harness} & 12.5 & 32.5 & 57.5 & 30.0 & 58.8 & 76.3 & 21.4 & 53.6 & 64.3 & 21.3 & 48.3 & 66.0 \\
AgenticRoboRL (Qwen3.5-9B SFT)~\cite{agenticroborl} & 10.0 & 27.5 & 50.0 & 5.0 & 21.3 & 41.3 & 3.6 & 21.4 & 46.4 & 6.2 & 23.4 & 45.9 \\
AgenticRoboRL (Qwen3.5-9B RL)~\cite{agenticroborl} & 5.0 & 27.5 & 55.0 & 7.5 & 28.8 & 51.3 & 0.0 & 17.9 & 46.4 & 4.2 & 24.7 & 50.9 \\
CaP-Agent0~\cite{capx} & 0.0 & 0.0 & 0.0 & 0.0 & 0.0 & 1.3 & 0.0 & 0.0 & 3.6 & 0.0 & 0.0 & 1.6 \\
ASPIRE~\cite{aspire} & 0.0 & 0.0 & 0.0 & 0.0 & 0.0 & 0.0 & 0.0 & 0.0 & 0.0 & 0.0 & 0.0 & 0.0 \\
RoboRSI~\cite{roborsi} & 0.0 & 0.0 & 5.0 & 0.0 & 0.0 & 2.5 & 0.0 & 0.0 & 0.0 & 0.0 & 0.0 & 2.5 \\
VIA~\cite{via} & 2.5 & 10.0 & 20.0 & 1.3 & 11.3 & 18.8 & 3.6 & 3.6 & 14.3 & 2.4 & 8.3 & 17.7 \\
\midrule
\rowcolor{blue!8}
\textbf{RA-RT (DeepSeek)} & \textbf{95.0} & \textbf{100.0} & \textbf{100.0} & \textbf{62.5} & \textbf{76.3} & \textbf{80.0} & \textbf{82.1} & \textbf{96.4} & \textbf{100.0} & \cellcolor{blue!8}\textbf{79.9} & \cellcolor{blue!8}\textbf{90.9} & \textbf{93.3} \\
\rowcolor{orange!10}
\textbf{RA-RT (Qwen3.5-9B)} & 90.0 & 92.5 & 97.5 & 48.8 & 57.5 & 72.5 & \textbf{82.1} & 89.3 & 96.4 & \cellcolor{orange!10}73.6 & \cellcolor{orange!10}79.8 & 88.8 \\
\bottomrule
\end{tabular}
\end{adjustbox}
\end{table}

\begin{table}[t]
\centering
\footnotesize
\setlength{\tabcolsep}{7pt}
\renewcommand{\arraystretch}{1.1}
\caption{\textbf{Scheduling ablation (top) and student progression (bottom).} Three-benchmark mean (\%), all-task \textbf{Med.\,(s)}, and \textbf{Cap.\,(s)} $= \mathbb{E}[\min(T_{\pi}, 1{,}800)]$. $^\dagger$$600$\,s cap without notes; $^\ddagger$full-$1{,}800$\,s vanilla serial run ($63.3\%$ / $380.7$\,s Med.\ at $600$\,s$^\dagger$).}
\label{tab:ablation_results}
\begin{adjustbox}{max width=\linewidth}
\begin{tabular}{l*{6}{c}}
\toprule
\textbf{Variant} & \textbf{200\,s} & \textbf{400\,s} & \textbf{800\,s} & \textbf{1800\,s} & \textbf{Med.} & \textbf{Cap.} \\
\midrule
\multicolumn{7}{@{}l}{\emph{Scheduling \& runtime (DeepSeek-V4.1-Flash)}} \\
Vanilla serial (RPent~\cite{harness})     & 29.5$^\dagger$ & 57.9$^\dagger$ & 63.3$^\dagger$ & 77.1$^\ddagger$ & 702.1$^\ddagger$ & 917.7$^\ddagger$ \\
+ Execution safeguards (serial)$^\dagger$ & 57.9 & 75.4 & 80.2 & 80.2 & 158.5 & 512.8 \\
+ Asynchronous scheduling$^\dagger$       & 79.7 & 86.0 & 86.8 & 86.8 & 60.0  & 352.5 \\
\rowcolor{blue!8}
\textbf{+ Restarts \& notes = RA-RT (DeepSeek)} & \textbf{79.9} & \textbf{90.9} & \textbf{93.3} & \cellcolor{blue!8}\textbf{93.3} & 59.5 & \textbf{253.2} \\
\midrule
\multicolumn{7}{@{}l}{\emph{Learning and inference (Qwen3.5-9B)}} \\
SFT student + 6-step MTP                  & 26.0 & 39.0 & 53.8 & 71.0 & 596.0 & 843.4 \\
+ Safeguard fixes \& action-only distill. & 34.9 & 49.2 & 69.1 & 78.6 & 408.2 & 700.5 \\
+ Aligned MTP \& 8-bit serving            & 56.5 & 65.5 & 76.1 & 82.3 & 187.9 & 581.8 \\
+ Time-aligned DAgger \& weight merging   & 60.0 & 71.0 & 82.3 & 83.5 & 132.0 & 489.1 \\
\rowcolor{orange!10}
\textbf{+ Refinement \& safeguards = RA-RT (9B)} & \textbf{73.6} & \textbf{79.8} & \textbf{88.8} & \cellcolor{orange!10}\textbf{92.1} & \textbf{46.7} & \textbf{343.8} \\
\bottomrule
\end{tabular}
\end{adjustbox}
\end{table}

\noindent\textbf{Effect of Asynchronous Scheduling.}
In \Cref{tab:ablation_results}, $^\dagger$ marks DeepSeek-V4.1-Flash runs without task notes, capped at 600 s and 100 turns per task. Later budget columns repeat the 600 s success rates. Safeguards raised mean serial completion from 63.3\% to 80.2\%. With safeguards fixed, asynchronous scheduling raised completion to 86.8\% and reduced the all-task median time from 158.5 s to 60.0 s. The separate 1,800 s serial run ($^\ddagger$, 360-turn limit) reached 77.1\%.
\paragraph{Analysis of Local Planner.}
Deployed locally without external APIs, RA-RT (Qwen3.5-9B) ranks second across all budgets in \cref{tab:benchmark_results}, outperforming RPent and both AgenticRoboRL models by wide margins. Even without notes, cold-start RA-RT (9B) reaches $85.9\%$ mean completion at $1{,}800$\,s ($\mathrm{Cap.}=435.1$\,s; supplement), beating $1{,}800$\,s vanilla serial DeepSeek ($77.1\%$, $917.7$\,s) in under half the expected time; with restarts and notes, it approaches RA-RT (DeepSeek) (\cref{fig:budget}), trailing mainly on Pro swaps.

\subsection{Efficiency Analysis}
\label{subsec:efficiency}

\begin{figure}[!htbp]
    \centering
    \begin{subfigure}[t]{0.48\linewidth}
        \centering
        \includegraphics[width=\linewidth]{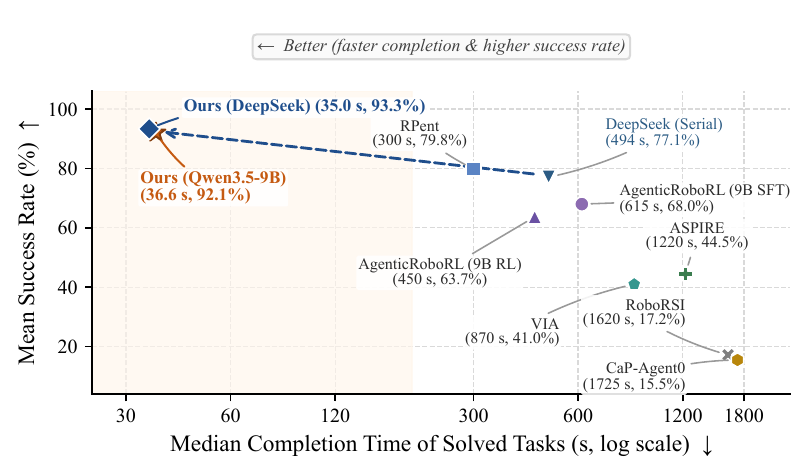}
        \caption{Mean success rate within 1{,}800\,s \vs median completion time of successful tasks.}
        \label{fig:tradeoff}
    \end{subfigure}
    \hfill
    \begin{subfigure}[t]{0.48\linewidth}
        \centering
        \includegraphics[width=\linewidth]{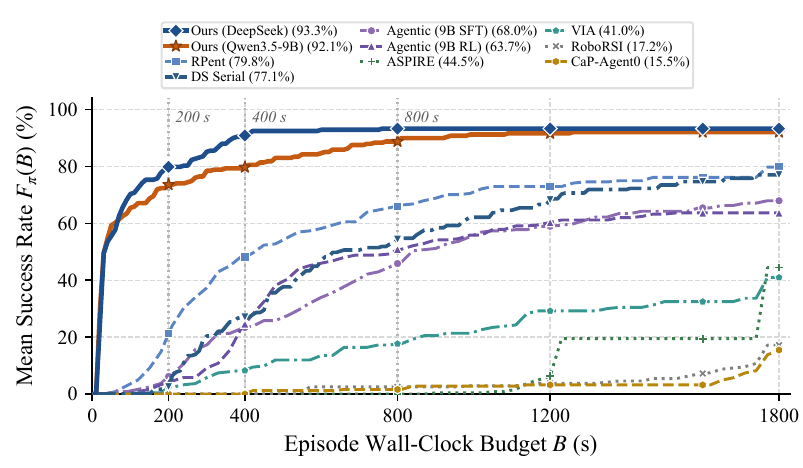}
        \caption{Mean success rate under different time budgets.}
        \label{fig:budget}
    \end{subfigure}
    \caption{\textbf{Time-constrained task completion and execution efficiency. } \textbf{(a)} Both of our planners sit in the upper-left high-completion, low-latency region, well ahead of blocking baselines. \textbf{(b)} Cumulative completion $F_{\pi}(B)$ over $B \in [0, 1{,}800]$\,s.}
    \label{fig:efficiency_analysis}
\end{figure}

By $200$\,s (\cref{fig:budget}), RA-RT (9B) matches the completion that blocking RPent needs $>20$\,min to reach. On solved tasks (\cref{fig:tradeoff}), RA-RT (DeepSeek) and RA-RT (9B) share similar medians ($35.0$\,s \vs $36.6$\,s); across all 148 tasks, the 9B planner has a lower median completion time (46.7 vs. 59.5 s), while the DeepSeek planner has a lower mean capped time (253.2 vs. 343.8 s).

\subsection{Ablation Study}
\label{subsec:ablation}

\Cref{tab:ablation_results} (bottom) tracks cumulative student gains (single-factor ablations in the supplement).
(1)~\emph{Speculative decoding and quantization.} On a paired $70$-task subset, 6-step MTP doubles decoding speed and cuts timeouts, while KL-aligned drafting and 8-bit serving halve median time. On our $24$-task validation set, 8-bit \texttt{W8A16}/\texttt{FP8} preserve accuracy, whereas 4-bit \texttt{NVFP4} suffers 3D drift (supplement).
(2)~\emph{Action-only supervision and DAgger.} Masking teacher \texttt{<think>} tokens (\cref{eq:token_weights}) with safeguard fixes lifts unmerged completion without verbose CoT drift (supplement); time-aligned DAgger with weight merging boosts long-budget recovery.
(3)~\emph{Multi-skill refinement and merging.} Short multi-skill rollouts accelerate early turns at the cost of hard recoveries (supplement), whereas merging with safeguards reduces capped mean time by a factor of 2.45 and median time by a factor of more than 12 relative to SFT.

\subsection{Real-World Deployment}
\label{subsec:real_world}

\begin{table}[!htb]
\centering
\caption{\textbf{Zero-shot real-world evaluation across four robot setups and nine manipulation tasks} ($10$ trials per multi-step pick-and-place task \textbf{P\&P}, $5$ trials per knob or box-folding task; $60$ trials total). Cells report successful trials, \textbf{Overall} success, and single-turn \textbf{TTFT} (local 9B on RTX PRO 5000 \vs the DeepSeek API); RPent~\cite{harness} supports Franka only.}
\label{tab:real_world_results}
\footnotesize
\setlength{\tabcolsep}{3.5pt}
\renewcommand{\arraystretch}{1.1}
\begin{adjustbox}{max width=\linewidth}
\begin{tabular}{l*{11}{c}}
\toprule
\multirow{2}{*}{\textbf{Method}} & \textbf{TTFT} & \multicolumn{2}{c}{\textbf{Franka (1-Arm)}} & \multicolumn{2}{c}{\textbf{RealMan (1-Arm)}} & \multicolumn{3}{c}{\textbf{ReBot (1-Arm)}} & \multicolumn{2}{c}{\textbf{RealMan (2-Arm)}} & \textbf{Overall} \\
\cmidrule(lr){3-4}\cmidrule(lr){5-6}\cmidrule(lr){7-9}\cmidrule(lr){10-11}
 & \textbf{(ms)} & P\&P & Knob & P\&P & Knob & P\&P & Knob & Box & 2-Knob & Box & \textbf{($n=60$)} \\
\midrule
RPent (DeepSeek)~\cite{harness}                & 1518 & 3/10 & 2/5 & --- & --- & --- & --- & --- & --- & --- & 5/15 (33.3\%) \\
Inspect Robots (DeepSeek)~\cite{inspectrobots} & 1200 & 2/10 & 3/5 & 3/10 & 1/5 & 1/10 & 2/5 & 0/5 & 1/5 & 0/5 & 13/60 (21.7\%) \\
AgenticRoboRL (9B RL)~\cite{agenticroborl}     & 1932 & 3/10 & 0/5 & 1/10 & 1/5 & 4/10 & 3/5 & 0/5 & 0/5 & 0/5 & 12/60 (20.0\%) \\
\midrule
DeepSeek-V4.1-Flash (Serial)                   & 1385 & \textbf{8/10} & 4/5 & 3/10 & 2/5 & 7/10 & \textbf{5/5} & 1/5 & 0/5 & 1/5 & 31/60 (51.7\%) \\
\rowcolor{blue!8}
\textbf{RA-RT (DeepSeek, Async)}                & 938 & \textbf{8/10} & \textbf{5/5} & \textbf{7/10} & 4/5 & \textbf{10/10} & \textbf{5/5} & \textbf{4/5} & 2/5 & \textbf{3/5} & \textbf{48/60 (80.0\%)} \\
RA-RT (Qwen3.5-9B, Serial)                      & 393 & 3/10 & 4/5 & 4/10 & 4/5 & 4/10 & 4/5 & 0/5 & 2/5 & 0/5 & 25/60 (41.7\%) \\
\rowcolor{orange!10}
\textbf{RA-RT (Qwen3.5-9B, Async)}              & 395 & 7/10 & 4/5 & 5/10 & \textbf{5/5} & 7/10 & \textbf{5/5} & 1/5 & \textbf{3/5} & 0/5 & 37/60 (61.7\%) \\
\bottomrule
\end{tabular}
\end{adjustbox}
\end{table}

\begin{figure}[!htb]
    \centering
    \includegraphics[width=\textwidth]{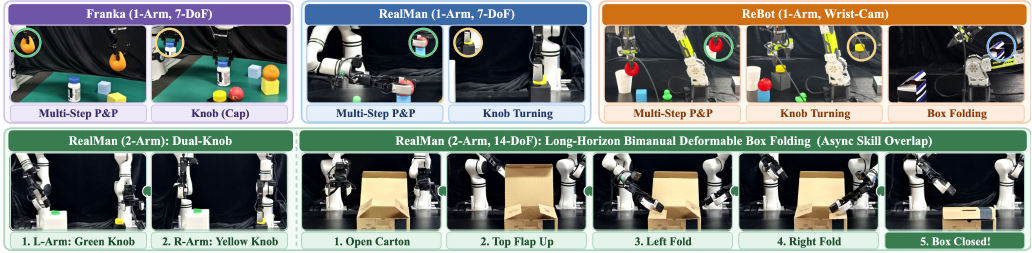}
    \caption{\textbf{Zero-shot real-world deployment across four setups and nine tasks.} Top: single-arm P\&P, knob/cap turning, and box folding on Franka, RealMan, and wrist-camera ReBot. Bottom: dual-knob turning and multi-stage deformable box folding on bimanual RealMan.}
    \label{fig:real_world}
\end{figure}

\noindent\textbf{Matched API Planner.}
Holding DeepSeek-V4.1-Flash fixed, RA-RT (DeepSeek, Async) outperforms its Serial counterpart and existing API frameworks (Inspect Robots~\cite{inspectrobots}, Franka-only RPent~\cite{harness}; \cref{fig:real_world,tab:real_world_results}) while lowering TTFT via compact context management. Rather than stalling after each read or motion, overlapping planning with execution queues predictable follow-up actions via \texttt{run\_if} and reacts early to slip or alignment errors, especially on wrist-camera ReBot and bimanual RealMan.
\par\noindent\textbf{Zero-Shot 9B Capability and Speed.}
Within the 9B class, RA-RT (Qwen3.5-9B, Async) far outperforms AgenticRoboRL (9B RL)~\cite{agenticroborl}, which is trained for blocking simulation calls and adapts poorly to physical skills. Despite being nearly two orders of magnitude smaller than the DeepSeek teacher~\cite{deepseekv41flash} and using \emph{zero} real-robot data, our local 9B planner transfers zero-shot across all four setups at sub-$0.4$\,s TTFT on one RTX PRO 5000. With asynchronous scheduling (whereas Serial mode drops completion), it beats cloud DeepSeek (Serial) overall, matches or edges RA-RT (DeepSeek, Async) on single- and dual-arm knobs, and transfers reliably to multi-step P\&P---combining generality and speed at 9B scale.
\par\noindent\textbf{Long-Horizon Deformable Tasks.}
Conversely, our 9B planner struggles on complex long-horizon tasks such as deformable box folding, where RA-RT (DeepSeek, Async) succeeds via broad pretraining priors. We trace this gap to two causes: (i)~\emph{bounded context}, as multi-stage folding tracks flap states over long histories where our $64$K cap and sliding visual/log window drop early cues; and (ii)~\emph{scarce data coverage}, as such tasks are nearly absent from our SFT, DAgger, and refinement datasets, making useful folding strategies hard to learn.

\section{Conclusion and Limitations}
\label{sec:conclusion}

We present an asynchronous skill-scheduling method that allows robots to plan subsequent steps while executing actions and adjust to execution feedback. Experiments show that our method improves task success rates and reduces completion time. Our local 9B planner outperforms the serial teacher, approaches the asynchronous teacher, and transfers to single- and dual-arm robots without fine-tuning on real-robot data. Three limitations remain. First, our 3D perception relies on calibrated depth sensors and requires a fallback to VLA skills when reflections cause missing depth measurements. Second, motion is not smoothly connected across skills, so the robot arm still pauses at skill boundaries. Third, although the simulation-trained planner transfers to real-world pick-and-place and knob-turning tasks, its generalization to deformable-object tasks such as box folding remains limited by insufficient simulation coverage.
Future work could test whether blending consecutive motions reduces pauses when the planner has already committed the next skill. For box folding, simulated cases in which a flap slips or reopens could help the planner learn when to repeat an earlier step.

\clearpage
\section*{LLM Usage Statement}
We used general-purpose large language models (Gemini and GPT models) solely as assistive tools during manuscript preparation and experiment development, specifically for language polishing, grammar refinement, LaTeX formatting, and routine code debugging. No LLMs were used to generate experimental results, fabricate citations, or formulate scientific claims. All text, mathematical derivations, code, and empirical findings in this paper were originally drafted, thoroughly verified, and approved by the authors, who take full responsibility for the entire submission.

\bibliographystyle{unsrtnat}
{\small
\bibliography{references}
}
\end{document}